%% file: eeav.tex
\documentclass[letterpaper, 10 pt, conference]{ieeeconf}

\usepackage{amsmath,amssymb,breqn}

\usepackage{algorithm}
\usepackage{algpseudocode}
\usepackage{xcolor} 
\usepackage{multirow} 
\definecolor{darkgreen}{RGB}{50, 150, 100} 

\usepackage{lipsum}
\usepackage[normalem]{ulem}
\useunder{\uline}{\ul}{}
\usepackage{makecell}
\usepackage{color}
\usepackage{graphicx}
\usepackage{subcaption}
\usepackage[    colorlinks=true,
    linkcolor=blue,
    filecolor=blue,      
    urlcolor=blue,
    citecolor=blue,]{hyperref}
\usepackage{hyperref}

\usepackage[english]{babel}
\newtheorem{remark}{Remark}
\usepackage{array}
\usepackage{booktabs}

\usepackage{colortbl} 

\IEEEoverridecommandlockouts                            

\title{\LARGE \bf
EMATO: Energy-Model-Aware Trajectory Optimization for Autonomous Driving
}

 \author{
     Zhaofeng Tian\(^{1}\), Lichen Xia\(^{1}\), and Weisong Shi\(^{1}\)
     \thanks{} 
     \thanks{\(^{1}\)The CAR Lab, University of Delaware, Newark, USA
             {\tt\small \{zhaofeng, lxia, weisong\}@udel.edu}} 
}

\begin{document}

\maketitle
\thispagestyle{empty}
\pagestyle{empty}

\begin{abstract}

Autonomous driving lacks strong proof of energy efficiency with the energy-model-agnostic trajectory planning. To achieve an energy consumption model-aware trajectory planning for autonomous driving, this study proposes an online nonlinear programming method that optimizes the polynomial trajectories generated by the Frenet polynomial method while considering both traffic trajectories and road slope prediction. This study further investigates how the energy model can be leveraged in different driving conditions to achieve higher energy efficiency. Case studies, quantitative studies, and ablation studies are conducted in a sedan and truck model to prove the effectiveness of the method.


\end{abstract}



\section{Introduction}


The energy consumption and associated economic impacts on the Autonomous Driving (AD) industry are significant. In 2021, US truck transportation covered 327.48 billion miles ~\cite{truck_mileage}, with fuel costs comprising approximately 30\% of the Total Cost of Ownership (TCO) ~\cite{liangkai_fuel, fuel_30}. While current AD trajectory planning research has achieved fundamental safe operations, substantial evidence of energy savings within an energy model-agnostic framework remains limited. This study proposes a novel \textbf{Energy Model Aware Trajectory Optimization (EMATO)} paradigm, enhancing energy efficiency through online nonlinear trajectory optimization on a differentiable energy model.

Conventional AD trajectory generation methods, such as cubic or polynomial curve generation, prioritize collision avoidance and passenger comfort through acceleration or jerk minimization~\cite{darpa_motion,state_lattice, poly_trajectory, frenet}. In these energy model-agnostic approaches, energy savings are often considered a byproduct of trajectory smoothing on acceleration~(general energy) rather than an explicit energy-model-based objective.


While applicable to various vehicle types, general energy representations lack robust proof of energy optimality. In contrast, Ecological Driving (ECO-driving) operates vehicles within high-efficiency zones, offering insights for energy-efficient autonomous driving through precise, model-based trajectory optimization. To explore the potential of integrating ECO-driving strategies, we examine related works.

 \begin{figure}[htbp]
    \centering
    \includegraphics[width=\linewidth]{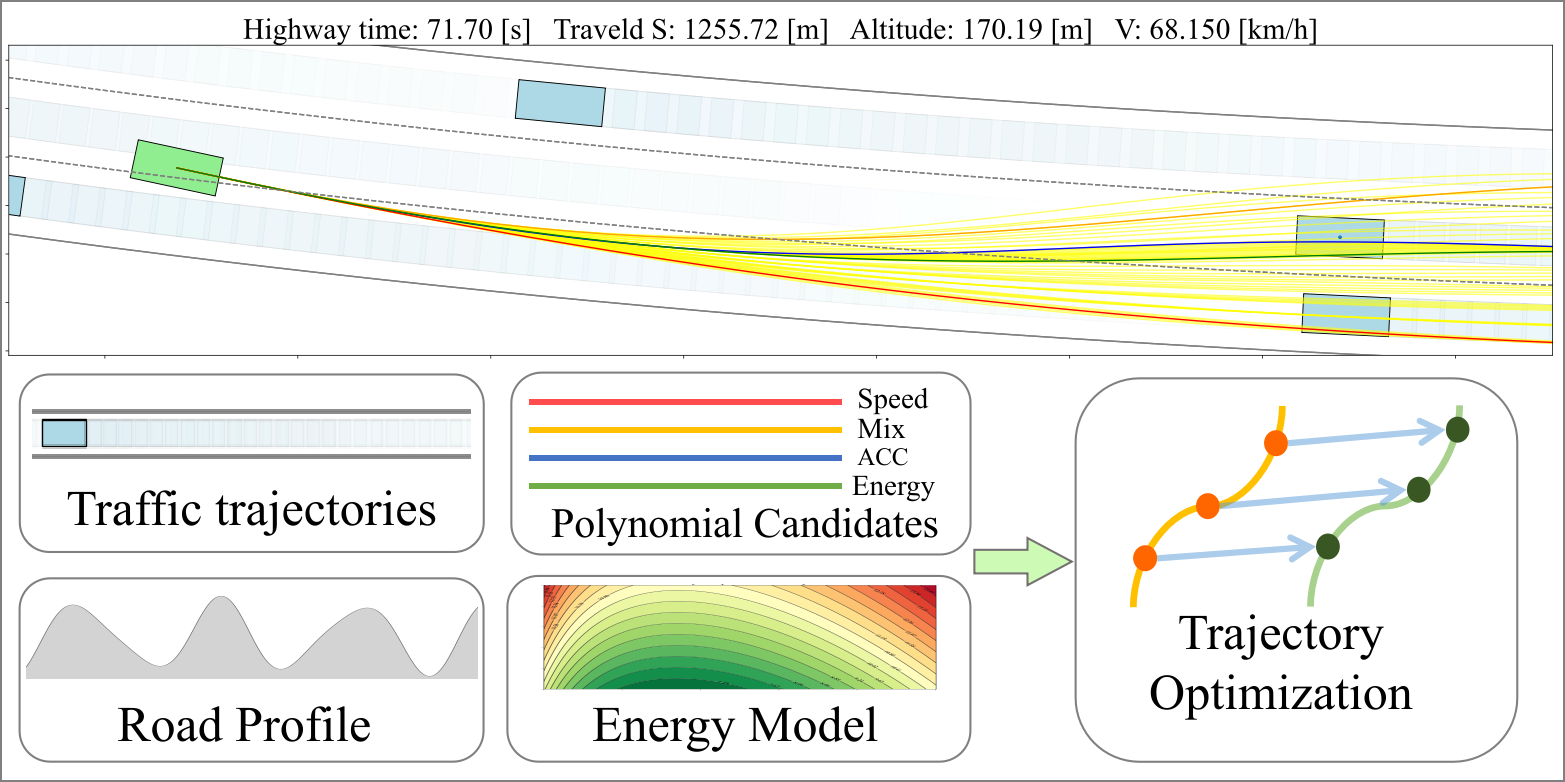} 
    \caption{EMATO framework in a Frenet highway system, it considers traffic and slope predictions and optimizes the sampled polynomial candidates with a differentiable energy model to achieve energy improvement.}
    \label{fig:frenet_frame}
\end{figure}

Pulse and Glide (PnG) strategies allow vehicles to operate at high-efficiency points, outperforming constant speed driving\cite{cao_png, png_shieh}. Study\cite{shengbo_png} converts PnG into a discrete Optimal Control Problem (OCP), considering engine speed, torque, and gear ratio. For Electric Vehicles (EV), genetic algorithms have been applied for offline PnG optimization~\cite{tian_png}, however, real-time application remains a challenge.

Eco-driving optimizes speed profiles given driving cycles and terrain, improving energy efficiency. Studies~\cite{lookahead_dp1, lookahead_dp2} use dynamic programming (DP) to optimize speed profiles and gear selection over elevation horizons, achieving 3.5\% fuel reduction without increasing trip time. Studies~\cite{down_slope,rolling_eco} use approximate bivariate polynomial energy functions and solve OCPs using Pontryagin's Maximum Principle~(PMP).

While ECO-driving strategies offer rich methodologies for energy-efficient AD, gaps exist in adapting these techniques to real-world applications. Generalizing ECO-driving from 1D to 2D scenarios (e.g., Frenet traffic-road systems) remains challenging, as current AD frameworks like Apollo's EM motion planner\cite{apollo_em} and Autoware's path sampler\cite{autoware_path} use state sampling and bounded value polynomial interpolation. Moreover, online solvability proof is crucial for real-time, safe autonomous driving applications~\cite{sidi_vehicle_computing1}. Showing merits in solving non-linear trajectory planning problems in a reasonable time~\cite{libai1,libai2,teb_rosmann,nlp_orca}, Non-Linear Programming (NLP) is leveraged in the proposed EMATO framework for the non-linear-energy-model-based optimization problem.


To address the research gaps, EMATO has the following contributions. \emph{\textbf{(i) Differentiable energy model}}: Adopting the model-based approach from ECO-driving, we innovate an accurate differentiable energy model concerning vehicle speed and attraction acceleration. \emph{\textbf{(ii) Frenet energy optimization}}: Leveraging the energy model, EMATO optimizes an energy-saving homotopic trajectory to the polynomial candidates with an online solving capability. \emph{\textbf{(iii) Extensive studies}}: We additionally validate the effectiveness of EMATO in Cruise Control (CC), PnG, and Adaptive Cruise Control (ACC) driving conditions for energy efficiency improvement.

The rest of the paper is arranged as follows. The energy model is fitted numerically in Sec~\ref{sec:energy_model}, EMATO case studies of PnG, ACC, and Frenet driving are illustrated in Sec~\ref{sec:emato}. While quantitative and ablation studies are conducted in Sec~\ref{sec:experiments}, and a conclusion drawn in Sec~\ref{sec:conclusion}.


\input{sec2}
\input{sec3}

\input{sec4}

\input{conclusion}

\bibliographystyle{IEEEtran}
\bibliography{main}

\end{document}

%% file: sec2.tex
\section{Differentiable Energy Model}
\label{sec:energy_model}

\subsection{Vehicle Dynamics}

Some necessary vehicle dynamic and fuel consumption formulas are introduced below for later fuel rate modeling.
    


\begin{figure}

    \centering
    \includegraphics[width=\linewidth]{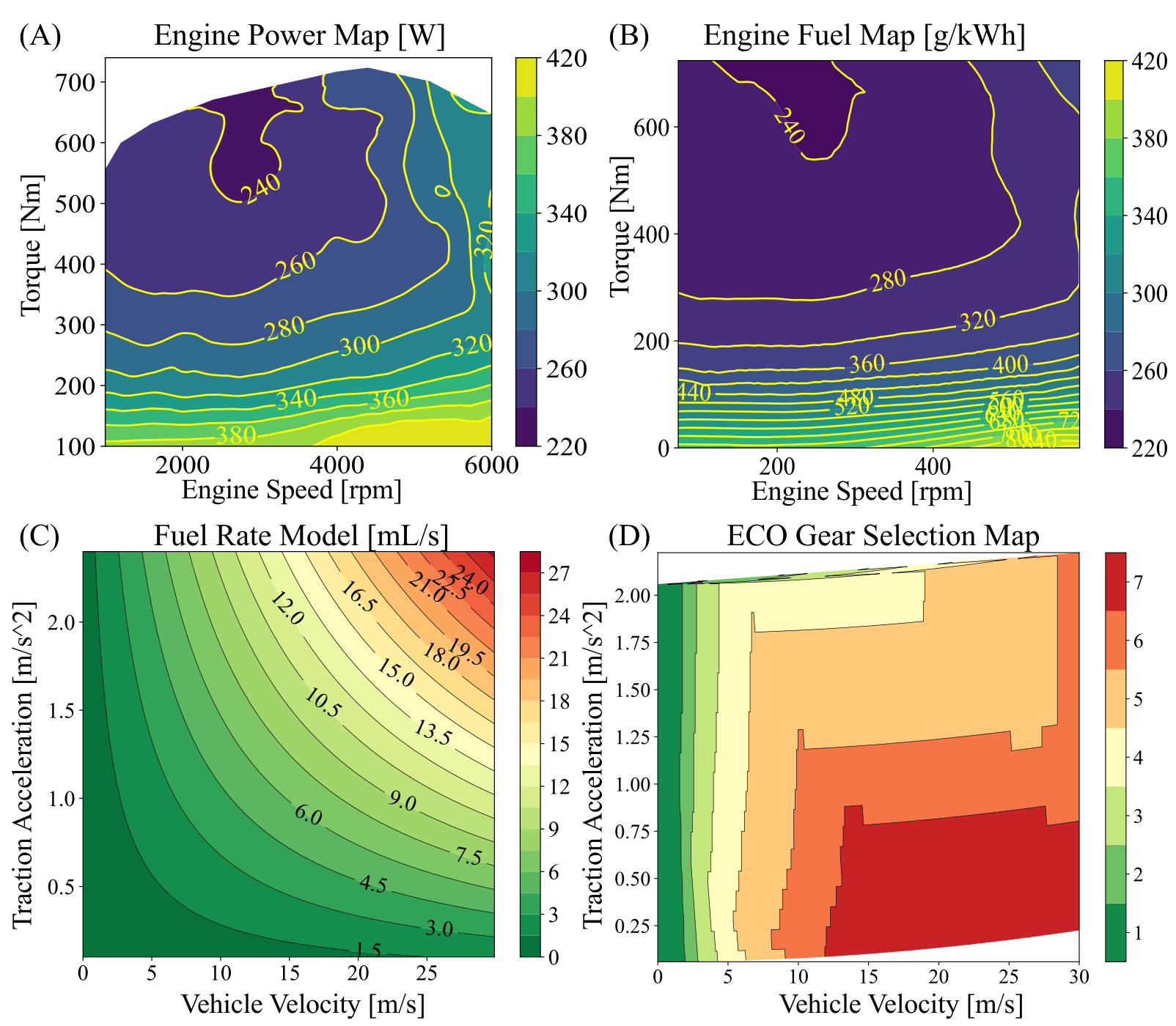}  
    \caption{(A-B) Original engine power and fuel map. (C-D) Fitted differentiable energy model and optimized gear selection w.r.t. traction acceleration $a_t$ and vehicle speed $v$ for a light-duty 7-speed truck.}
\label{fig:map}
\end{figure}






\begin{equation}
\label{eq_ft}
    F_t = a_t \cdot M, \quad T_w = F_t \cdot r
\end{equation}
\begin{equation}
    \label{eq_te}
      T_e = \frac{T_w }{i_t \cdot \eta},  \quad \omega_e = \frac{v \cdot i_t }{r}, \quad i_t = i_g \cdot i_f  
\end{equation}
\begin{equation}
\label{eq_pe}
    P_e = \mathcal{F}_P(\omega_e, T_e), \quad \eta_{f} = \mathcal{F}_{fe}(\omega_e, T_e)   
\end{equation}
    
\begin{equation}
\label{eq_fr}
    f_r = \frac{P_e \cdot \eta_f}{c_u} 
\end{equation}

\begin{equation}
    \label{eq:k}
    k_1 = \frac{C_d \cdot \rho \cdot A_v}{2 \cdot M}, \quad k_2 = \mu \cdot g , \quad k_3 = g
\end{equation}

\begin{equation}
    \label{eq:ar}
    a_r = k_1 \cdot v^2 + k_2 \cdot cos(\theta(s)) + k_3 \cdot sin(\theta(s))
\end{equation}

\begin{equation}
\label{eq:at}
    \underbrace{a_t = a_v +  a_r + a_b}_{\text{Vehicle Dynamics}}
\end{equation}

\begin{equation}
\label{eq:transit}
\underbrace{\begin{pmatrix}
\Delta l \\
\Delta v \\
\Delta a_v
\end{pmatrix}
=
\int_{t_0}^{t_k}
\begin{pmatrix}
0 & 1 & 0 \\
0 & 0 & 1 \\
0 & 0 & 0
\end{pmatrix}
\begin{pmatrix}
l(\tau) \\
v(\tau) \\
a_v(\tau)
\end{pmatrix}
\, d\tau
+
\int_{t_0}^{t_k}
\begin{pmatrix}
0 \\
0 \\
1
\end{pmatrix}
j(\tau) \, d\tau}_{\text{State Transition}}
\end{equation}

Where $C_d$, $\rho$, $A_v$, and $M$ respectively denote air drag coefficient, air density, frontal area, and equivalent mass; $\mu$ and $g$ are the rolling resistance coefficient and gravity value; $l$, $v$, $a_t$, $j$ are the vehicle's path distance, velocity, traction acceleration and jerk; $a_v$, $a_r$, $a_b$ denote apparent acceleration, resistance acceleration, and brake acceleration; Road slope/grade is denoted by $\theta(s)$, which generally can be obtained from road profiling or GPS information from road coordinate $s$~\cite{down_slope, rolling_eco}. 

To acquire the fuel rate, referring to Eqs.~(\ref{eq_ft}-\ref{eq_fr}), engine torque $T_e$ can be derived from wheel torque $T_w$ with traction force $F_t$,  wheel radius $r$, the transmission ratio $i_t$, and efficiency $\eta_t$, in which $i_t$ equals to the product of gear ratio $i_g$ and final drive ratio $i_f$, where traction force $F_t$ is directly related to traction acceleration $a_t$; Engine speed $\omega_e$ can be calculated with vehicle speed $v$ and transmission ratio $i_t$; By using two interpolated functions $\mathcal{F}_P$ and $\mathcal{F}_{fe}$, engine power $P_e$(w) and fuel efficiency $\eta_f$ (g/kwh) can be aquired from $\omega_e$, $T_e$ as shown in Fig.~\ref{fig:map}(A)(B), where the truck data is from the professional truck simulator TruckSim~\cite{trucksim}; A fuel rate $f_r$ at unit (mL/s) referring to study~\cite{down_slope}, can be derived from the product of $P_e$ and $\eta_f$ with $c_u = \rho_g \cdot 1000 \cdot 3600$ that converts the consumption rate unit to (mL/s), where $\rho_g = 0.85$ (g/mL) represents the disel density.

\subsection{Fuel Rate Modeling}
Previous works tend to approximate the fuel consumption map into a bivariate quadralic~\cite{va_fuel_model} or polynomial~\cite{down_slope, down_slope_traffic} functions with vehicle speed $v$ and apparent acceleration $a_v$, achieving a differentiable numerical optimization. Considering the traction acceleration $a_t$ would involve fuel consumption in a more direct way than apparent acceleration $a_v$ as slope $\theta(\cdot)$ is also incorporated referring to Eqs.~(\ref{eq:k}-\ref{eq:at}), we optimize the gear selection and fit the energy model w.r.t. $v$ and $a_t$. With a similar fitting function in~\cite{down_slope} with $a_v$ replaced by $a_t$ in Equation~(\ref{eq:fit}).
\begin{equation}
    \label{eq:fit}
     \hat{f_r}(v, u) = o_0 + o_1v +o_2v^2 + o_3v^3 + o_4v^4 + (c_0+c_1v+c_2v^2)a_t
\end{equation}

To eliminate the discrete terms in an NLP setting, an ECO gear selection policy is offline-optimized with the engine data that enables the engine to work on high-efficiency zones~\cite{shengbo_png}. With this ECO gear selection policy and $v$, $a_t$ samples in a dataset, we design a min-square objective function below that minimizes the squared error over the estimated fuel rate $\hat{f_r}^{i}$ and the real fuel rate $f_r^{i}$ based on engine maps and Eqs.~(\ref{eq_ft}-\ref{eq_fr}), so that fits the model parameters $\mathbf{o}$ and $\mathbf{c}$:
\begin{equation}
J_{fit} = \min_{o_0, o_1, o_2, o_3, o_4, c_0, c_1, c_2} \sum_{i=1}^{N} \left( f_r^{i} - \hat{f_r}^{i} \right)^2
\end{equation}
The fitted differential fuel model and the optimized gear selection policy are illustrated in Fig.~\ref{fig:map}(C-D). The average prediction accuracy is 98.22\%, which is considered precise.

%% file: sec3.tex
\section{Energy Model Aware Trajectory Optimization}
\label{sec:emato}
In this section, we present EMATO, an online NLP-based trajectory optimization framework that optimizes energy efficiency. Then we investigate how EMATO can be applied to different driving conditions.

\subsection{Problem Formulation}

We define the physical workspace $\mathcal{W} \subseteq \mathbb{R}^2$ where the ego agent drives in. At a single time step, the kinematic state in EMATO can be defined as $ x = [l, v, a_v] \in \mathbb{R}^{3}$, the observation state as $z = [l, v, a_v, j, \theta, a_r, f_r] \in \mathbb{R}^7 $, and $x \subset z$, the control state as $ u = [a_t, a_b]\in \mathbb{R}^{2}$,  the interference space induced by traffic agents' predicted trajectories as $\mathcal{W}_{\text{int}}$.  Further, a trajectory over a time horizon $ t \in [t_0, T]$, with a discrete length $n_T = \frac{T-t_0}{dt} $, where $dt$ is the time resolution, can be denoted as bold signs, for example:
\begin{equation}
\begin{aligned}
    \label{eq:x}
    & \mathbf{z} = \{[l, v, a_{v}, j, \theta, a_r, f_r]_t \,|\, t \in [t_0, T]\} \in \mathbb{R}^{n_{T} \times 7}\\
    & \mathbf{z}(t) = z_t = [l, v, a_v, j, \theta, a_r, f_r]_t \in \mathbb{R}^7 
\end{aligned}
\end{equation}

\renewcommand{\Comment}[1]{\textcolor[rgb]{0.043, 0.635, 0.604}{// #1}}

\begin{algorithm}
\caption{EMATO Framework }
\label{alg:EMATO}
\begin{algorithmic}[1]
    \State \textbf{Parameters:} $\mathbf{z}_r$, $\mathbf{u}_r$, $\mathbf{p}_g$, $\mathcal{Z}_{a}$, $\mathbf{w}$, $f_z$, $g_{\text{init}}$, $g_{\text{end}}$, $g_{\text{itm}}$, $[\underline{g}, \overline{g}]$,
      $[t_0, T]$;
    \Statex \Comment{Variable initialization}
    \State $\mathbf{z}_0, \mathbf{u}_0 \gets \text{initial guess based on } \mathbf{z}_r, \mathbf{u}_r$;
    \State $\mathbf{p}_{g0} \gets \text{slope profile prediction based on }\mathbf{l}_r \in \mathbf{x}_r$;
    \Statex \Comment{State and dynamic constraints}

    \State Update $\mathcal{W}_{\text{int}}$ based on traffic predictions $\mathcal{Z}_{a}$;
    \State  $g_{\text{init}},g_{\text{end}} \gets$ start-end state constraints;
    \State $g_{\text{itm}} \gets$ vehicle limits and working conditions;
    \State $f_z \gets$ dynamics with $\mathbf{p}_{g0}$ in Eqs.~(\ref{eq:k}-\ref{eq:transit})~(\ref{eq:fit});
    \Statex \Comment{Optimization problem set up}
    \State Update objective $J$ with $\mathbf{w}$;
    \State Set up NLP problem~(\ref{eq:emato}) with ($\mathbf{z}_0, \mathbf{u}_0$,  $\mathbf{p}_{g0}$, $f_s$, $\mathcal{W}_{\text{int}}$,  $g_{\text{init}}$, $g_{\text{end}}$, $g_{\text{itm}}$, $[\underline{g}, \overline{g}]$, $[t_0, T]$, $J$);
    \State $\mathbf{z}, \mathbf{u} \gets \text{solve NLP problem}$;
    \State \Return $\mathbf{z}, \mathbf{u}$
\end{algorithmic}
\end{algorithm}

\newcommand{\mycomment}[1]{\hfill \textcolor{gray}{\(\triangleright\) #1}}

Despite the convenience of formulating the problem over the path coordinate $l$ in a 1D global optimization~\cite{ocp_scp}, formulation over the time horizon $T$ would better adapt to the highly dynamic 2D local environment, and support fast trajectory rollout. The EMATO problem can be abstracted numerically to a NLP problem as :

\begin{equation}
\begin{aligned}
    \label{eq:emato}
    \min_{(\mathbf{z}(t), \mathbf{u}(t))} \quad & J(\mathbf{z}(t), \mathbf{u}(t), [t_0, T]), \\
    \text{s.t.} \quad \quad & \dot{\mathbf{z}}(t) = f_z (\mathbf{z}(t), \mathbf{u}(t), \mathbf{p}_g) \\ & f_p(\mathbf{z}(t)) \not\subset \mathcal{W}_{\text{int}}, \, \forall t \in [t_0, T];\\
    & g_{\text{init}}(\mathbf{z}(t_0)) = 0, \\
    & \underline{g}_t \leq g_{\text{itm}}(\mathbf{z}, \mathbf{u}) \leq \overline{g}_t \\
        & \underline{g}_T \leq g_{\text{end}}(\mathbf{z}(T)) \leq \overline{g}_T \\
\end{aligned}
\end{equation}
 $f_z$ denotes the state transition function that covers the dynamics equations~(\ref{eq:k}-\ref{eq:fit}) on the differential energy model, where the parameter $\mathbf{p}_g $ is the road slope profile that $\theta(l) = \mathbf{p}_g(l)$; $g_{\text{init}}$, $g_{\text{end}}$, and $g_{\text{itm}}$ are constraint functions for initial, end, and intermediate states. $[\underline{g}_T, \overline{g}_T]$, $ [\underline{g}_t, \overline{g}_t]$ are the box boundaries for $g_{\text{end}}$ and $g_{\text{itm}}$; $[\underline{\mathbf{z}},\overline{\mathbf{z}}]$ and  $[\underline{\mathbf{u}},\overline{\mathbf{u}}]$ are trajectory-level lower and upper boundaries of $\mathbf{z}$ and $\mathbf{u}$ over the horizon $[t_0, T]$; A general EMATO objective function $J$ can be defined as:

\begin{equation}
    \label{eq:objective}
    J =  \int_{\tau_0}^{T} 
     \Big( \underbrace{\left\| v(\tau) - v_d, \, a_v(\tau), \,a_b(\tau), \, j(\tau)]\right\|_{2, \mathbf{w_g}}^2}_{\text{weighted squared general objectives}}  \underbrace{+\frac{w_f \cdot f_r(\tau)}{v(\tau)}}_{\text{fuel efficiency}}\Big)
    d\tau
\end{equation}

We slightly abuse the notion of $\mathbf{w_g} = [w_v, w_a, w_b, w_j] \in \mathbb{R}^4$ to denote weights applied to squared general objective components in the first term, where $v(\tau) - v_d$ is a desired speed tracking component, and $a_v$, $a_b$, $j$ components account for passenger comfort (they also impact energy performance, see ablation studies). The second term denotes fuel efficiency measured in (mL/m) with a weight $w_f \in \mathbb{R}^1$, which is optimized directly in the objectives and therefore achieves better energy efficiency. We set $\mathbf{w} = [\mathbf{w}_g, w_f] \in \mathbb{R}^5$ as the total weights for $J$.  The overall EMATO framework is illustrated in Alg.~\ref{alg:EMATO}, where it takes a reference trajectory as input to initially guess the unsolved $\mathbf{z}$ and $\mathbf{u}$ for a warm start; The predicted road slope profile $\mathbf{p}_{g0} = [\theta_{t_0},\theta_{t_1}, ... , \theta_{T}]$ can be determined by looking up the original profile $\mathbf{p}_g$ with reference path coordinate trajectory $\mathbf{l}_r \in \mathbf{z}_r$; $[\underline{g}, \overline{g}]$ contains the overall boundary values of all constraint functions. 

\begin{remark}
  Given the reference trajectory being the resolution of the last iteration in an iterative replanning framework, the error between $\mathbf{p}_{g0}$ and the real slope trajectory w.r.t. solved $\mathbf{l} \in \mathbf{z}$ in the current iteration is very small, fixing it at the initialization phase would circumvent reiterations.  
\end{remark}

\subsection{Cruise Control and PnG}
\label{sec:png}

CC or constant speed driving at a traffic flow or limit speed is intuitively considered comfortable and efficient for human drivers. Some studies have shown the PnG could have better energy performance than CC~\cite{png_shieh,cao_png,tian_png,shengbo_png}. In a 900-m 1D driving test at a desired speed $v_d = 20$m/s, and a replanning interval $dl = 150$m, we compare the commonly used quintic polynomial method~\cite{frenet}, CC, and EMATO policy.

\begin{figure}
    \centering
    \includegraphics[width=\linewidth]{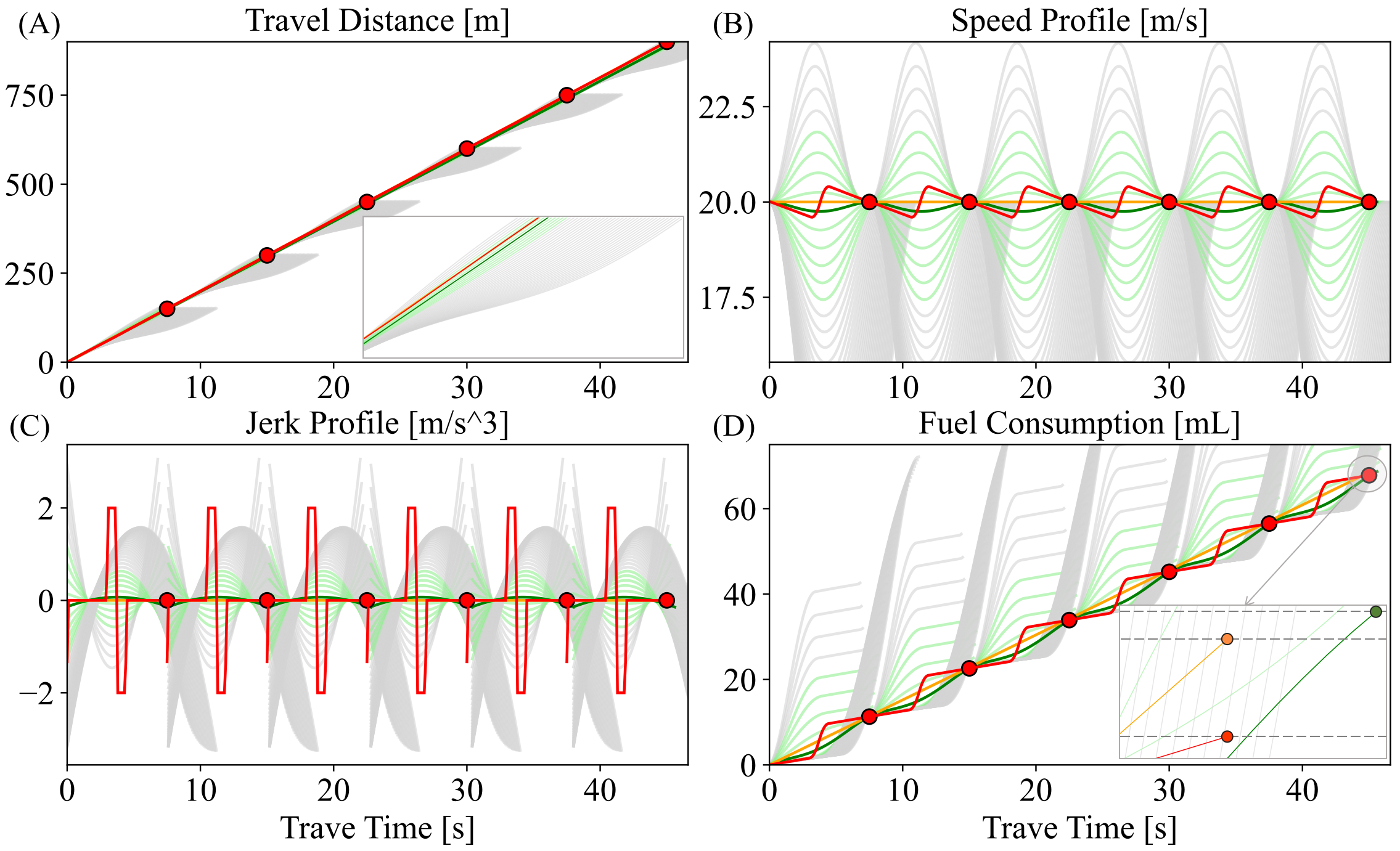}  
    \caption{Gray and light green lines denote overjerky and feasible quintic candidates. Green and red lines denote the ``Energy'' quintic candidate and the PnG trajectory produced by EMATO. }
\label{fig:png}
\end{figure}

In the quintic method, a polynomial trajectory is interpolated given a pair of start-end states within a sampling time window. Further, polynomial candidates are generated, and the overall observation and control trajectory sets are derived by dynamic equations and a road slope profile (set as ``Flat" in this case study). Using the same objective $J$ in Eq.(\ref{eq:objective}) with different weights, different driving policies can be set among the candidates. The "Energy" policy uses $[[0]_{1\times4},1]$ to achieve the best fuel efficiency, and the respective trajectory is plotted with the green line in Fig.\ref{fig:png}; 

The CC trajectory can also be obtained by the quintic method~(with the corresponding state sample), and for a single iteration, a Bounded Value Problem (BVP) with partial constraints is set up, using the CC trajectory as a reference. With a $\mathbf{w} = [0, [0.0001]_{1\times3},35]$, we have the energy--efficiency-optimized trajectory $\mathbf{z}_e, \mathbf{u}_e \gets  \textit{EMATO}( \mathbf{z}_{\text{cc}},\mathbf{u}_{\text{cc}},\mathbf{p}_g, \mathbf{w}, g_{\text{init}}, g_{\text{end}} | \, \cdot)$. The speed profile $\mathbf{v}_e$ in Fig.~\ref{fig:png}(B) shows the zig-zag contour matching the PnG features, and the generated PnG trajectory achieves lower consumption at 67.88 mL and 1.16\% efficiency improvement in miles per gallon~(mph)  compared to CC at 68.47~mL without temporal loss. ``Energy'' candidate consumes 68.67~mL. 

\begin{remark}
  Depending on sampling density and quality, the energy efficiency of the quintic method can approach that of CC. However, EMATO outperforms CC by producing a PnG trajectory, demonstrating its effectiveness.
\end{remark}

\begin{table}[htbp]
\caption{ACC Algorithm Library}
\centering
\begin{tabular}{m{1.21cm}<{\centering}|m{1.5cm}<{\centering}|m{1.5cm}<{\centering}|m{3cm}}
\Xhline{1.2pt}
\textbf{Algorithm ID}& \textbf{Intermediate $l_l- l$} & \textbf{End $l_l- l$ } & \textbf{Description} \\
\hline
 \textit{Quintic} & None  & $ [\Delta l_{\text{acc}}, \Delta l_{\text{max}}]$  &  ``Energy'' policy, BVP s.t. end state ACC constraint.\\  
\hline
 \textit{EMATO-B}& $ [\Delta l_s, \Delta l_{\text{max}}]$   & $ [\Delta l_{\text{acc}}, \Delta l_{\text{acc}}]$  &  BVP, end state tightly ACC-constrained. \\ 
\hline
 \textit{EMATO-R}& $[\Delta l_s, \Delta l_{\text{max}}]$  & \makecell[c]{$ [\Delta l_{\text{acc}}-\Delta l_r,$ \\ $\Delta l_{\text{acc}} + \Delta l_r]$} & End state ACC-constraint relaxed by $\Delta l_r$. \\ 
 \hline
 \textit{EMATO-V} & $[\Delta l_s, \Delta l_{\text{max}}]$ & $[\Delta l_s, \Delta l_{\text{max}}]$  & Constraints relaxed, tracking $v_l$ for ACC behavior. \\ 
\Xhline{1.2pt}
\end{tabular}

\label{tab:acc}
\end{table}

\subsection{Adaptive Cruise Control}
\begin{figure}
    \centering
    \includegraphics[width=\linewidth]{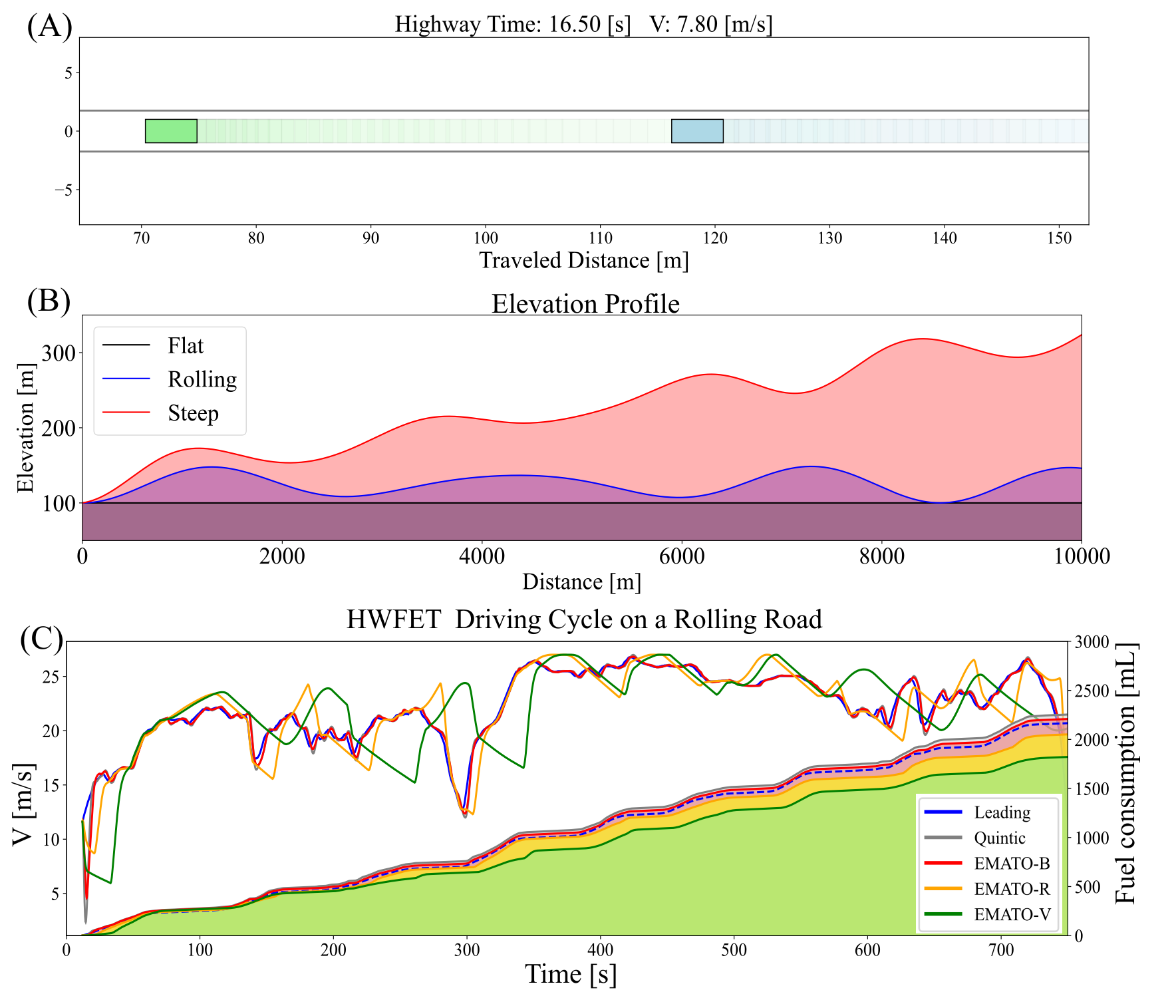}  
    \caption{(A) ACC simulation. (B) Simulated flat, rolling, and steep elevation profiles. (C) Case study results of quintic and EMATO methods in a HWFET cycle with a rolling road.}
\label{fig:acc}
\end{figure}

In this case study, we investigate how EMATO benefits energy efficiency during an ACC operation. We simulate a leading vehicle running a fuel test driving cycle, i.e., ``Highway Fuel Economy Test Cycle (HWFET)~\cite{hwfet}'', and the ego vehicle operating ACC to adaptively follow the leading vehicle. Besides ``Flat'', we generate two road slope profiles ``Rolling'' $\mathbf{p}_r$ and ``Steep'' $\mathbf{p}_s$ to simulate namely rolling and steep roads shown in Fig.~\ref{fig:acc}(B). A linear projection is applied on $\mathbf{x}_l(t)$ in the driving cycle profile to obtain the leading vehicle trajectory $ \mathbf{z}_l,\mathbf{u}_l $ over a prediction window $[t_0, T]$. A typical ACC spacing strategy~\cite{tian_png} at a single timestep can be written as:

\begin{equation}
    \label{eq:acc}
    \Delta l_{\text{acc}} = l_{l} -l  = T_h \cdot  v_{l} + \Delta l_s
\end{equation}

In study~\cite{frenet}, this spacing strategy is set as an end-state constraint in a quintic polynomial method, we refer to this setting for the quintic method and set EMATO algorithms for the ACC problem in Table~\ref{tab:acc}. Where \textit{EMATO-B} and \textit{EMATO-R} are respectively tightly bounded and relaxed bounded at end states with ACC spacing constraint, while in the intermediate states, only minimum safe distance should be kept from without velocity objective; \textit{EMATO-V} only constrain the minimum and maximum space with the leading car and instead operate ACC by tracking the leading car's speed. The weight setting for $J$ in the above methods is listed in ~\cite{emato_website}. Uniformly, the optimized trajectory can be acquired by $\mathbf{z}_e, \mathbf{u}_e \gets  \textit{EMATO}( \mathbf{z}_{l},\mathbf{u}_{l},\mathbf{p}_g, \mathbf{w}, g_{\text{itm}}, g_{\text{end}} | \, \cdot)$. With HWFET cycle and a rolling road, the comparison of four methods can be observed in Fig.~\ref{fig:acc}, where EMATO methods outperform a quintic method with lower fuel consumption. More quantative experiments are in Sec.~\ref{sec:experiments}.

\begin{remark}
The ability to track the leading vehicle's speed improves when the ACC end-state boundary is tighter in EMATO, but energy efficiency goes the opposite.
\end{remark}



\subsection{Frenet Autonomous Driving}
\begin{algorithm}
\caption{Iterative Frenet EMATO Framework}
\label{alg:frenet}
\begin{algorithmic}[1]
    \State \textbf{Initialize:} $\{ \mathbf{x}_s^i(t_0), \mathbf{x}_s^i(T) \}, \{ \mathbf{x}_d^i(t_0), \mathbf{x}_d^i(T) \}, f_z, \mathcal{Z}_a, \mathbf{P}_g $ 
    
    \For{each rollout}
        \Statex \, \quad \Comment{Generate Frenet polynomial candidates}
        \For{each $\left[\mathbf{x}_s^i(t_0), \mathbf{x}_s^i(T)\right]$ and $\left[\mathbf{x}_d^i(t_0), \mathbf{x}_d^i(T)\right]$}
            \State $\mathbf{x}_s^i, \mathbf{x}_d^i \gets \text{quintic-interpolate start-end state pair;} $
            \State $\mathbf{z}_p^i, \mathbf{u}_p^i \gets \text{transform } \mathbf{x}_s^i, \mathbf{x}_d^i $ to a path-coordinate \Statex \qquad \qquad \qquad \quad based trajectory with $\mathbf{P}_g$;
            \State Append $\mathbf{z}_p^i, \mathbf{u}_p^i$ to candidates set $\mathcal{C}$;
        \EndFor
        \State Check feasibility in $\mathcal{C}$ with traffic predictions $\mathcal{Z}_a$;
        \State $\mathbf{z}_p^o, \mathbf{u}_p^o \gets$ select a Frenet candidate with an objective.
        \Statex \, \quad \Comment{BVP with selected Frenet polynomial trajectory}
        \State Set $g_{\text{init}}, g_{\text{end}}$ with $\mathbf{z}_p^o$, tight bounds for $[\underline{g}_T, \overline{g}_T]$;
        \State $\mathbf{z},\mathbf{u} \gets \textit{EMATO}(\mathbf{z}_p^o, \mathbf{u}_p^o,\mathbf{p}_g, \mathcal{Z}_a ,f_z, g_{\text{init}}, g_{\text{end}}, [\underline{g}_T, \overline{g}_T]|\cdot )$
        \Statex \, \quad \Comment{State transition}
        \State Update new planning horizon $[t_0, T]$, vehicle states, \Statex \, \quad $[s, d]$ sampling set, $\mathcal{Z}_a$;  
    \EndFor
    
    \State \Return $\mathbf{z}, \mathbf{u}$
\end{algorithmic}
\end{algorithm}

\begin{figure}
    \centering
    \includegraphics[width=\linewidth]{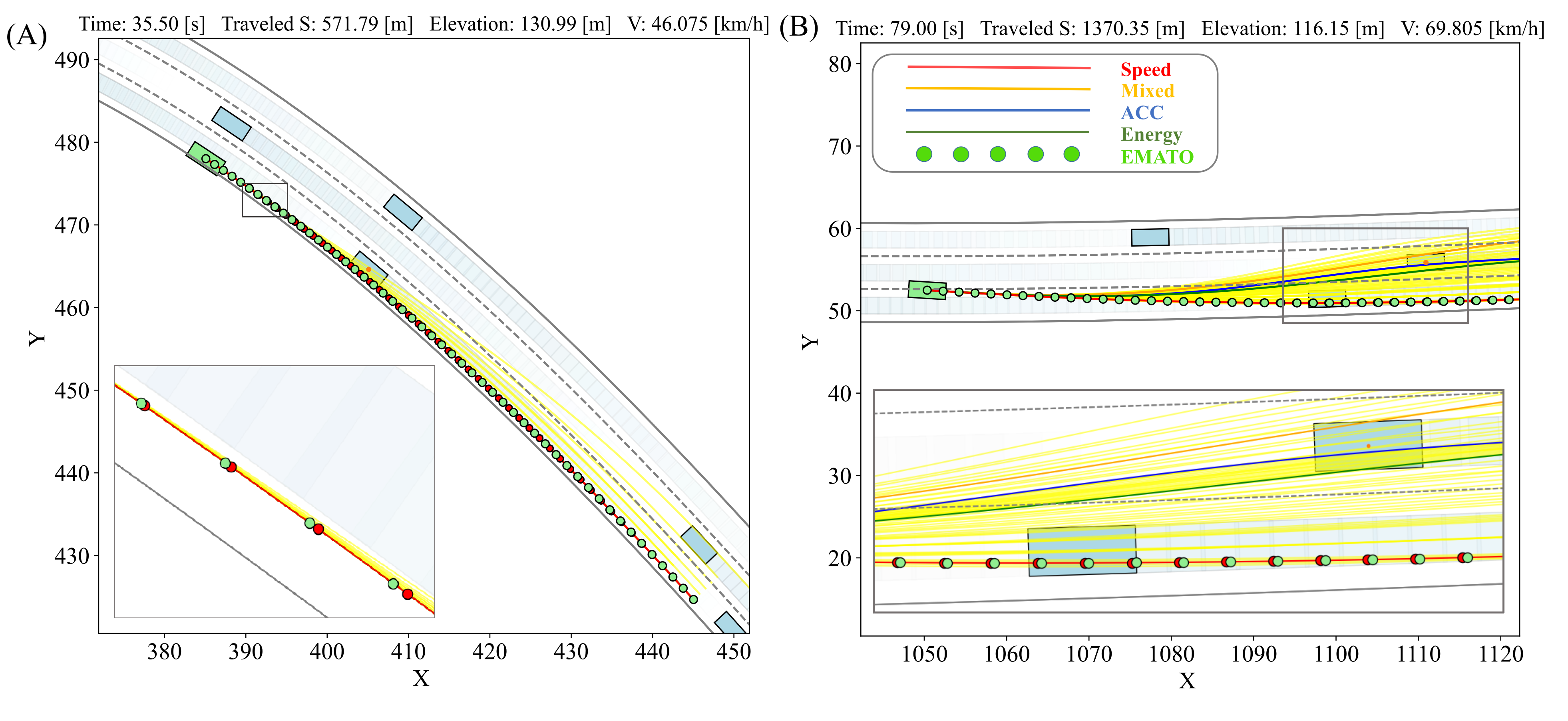}  
    \caption{Frenet highway simulation. (A-B) An example of applying EMATO to a ``Speed'' quintic trajectory (red), and the optimized trajectory waypoints are plotted in green dots.}
\label{fig:frenet}
\end{figure}

A Frenet polynomial trajectory can be interpolated with longitudinal start-end states $\mathbf{x}_s(t_0) = [s, s', s'']_{t_0}, \mathbf{x}_s(T) = [s, s', s'']_{T}$, and lateral states $\mathbf{x}_d(t_0) = [d, d', d'']_{t_0}, \mathbf{x}_d(T) = [d, d', d'']_{T}$ referring to study~\cite{frenet}. An abstracted Frenet EMATO framework is shown in Fig.~\ref{fig:frenet}. And the specific framework is expressed by Alg.\ref{alg:frenet}, where \textbf{in line~3-8}, Frenet candidates are interpolated with sampled start-end states for both s coordinate trajectory $\mathbf{x}_s$ and d coordinate $\mathbf{x}_d$ over a time horizon $[t_0, t_0 + \Delta t = T]$. $\Delta t$ is fixed for not changing the trajectory length for a faster problem update instead of rebuilding the problem every rollout in the NLP program; Given a [s,~d] trajectory, an overall trajectory in Global coordinate $[x, y, \text{yaw}]$ can be derivated with a Frenet-Global transformation~\cite{frenet}. Further, the trajectory on the path coordinate $\mathbf{z}_p, \mathbf{u}_p$, can be inferred by the trajectory on the Global coordinate. For consistency with previous problem settings, the extra global intermediate states in the Alg.~\ref{alg:frenet} are not shown; \textbf{In line~9}, among the valid quintic candidates (collision, overjerky, overcurvy candidates are infeasible), by different weights setting in $J$, different policies can be generated as Table.~\ref{tab:frenet}; \textbf{Line~10, 11} illustrate EMATO is applied to a quintic Frenet candidate in a BVP form~(same start-end constraints in Sec.~\ref{sec:png}), and the optimized trajectory is homotopic to the original trajectory without path shape change (time reallocation in path coordinate $l$), which is shown in Fig.~\ref{fig:frenet_frame}; $w_v$ in EMATO weights are set to 0 since quintic start-end states have already bounded it.  Moreover, the corresponding EMATO algorithms to the quintic methods are listed in Table.~\ref{tab:frenet}, and a quantative comparison is conducted in Sec.~\ref{sec:experiments}.

\begin{remark}
Integrating EMATO into a 2D quintic method preserves the quintic method's advantages of fast state sampling, candidate generation, and collision checking.
\end{remark}

\begin{remark}
EMATO homotopy is safety-guaranteed if the maximum waypoint distance change $\xi$ is within the safety boundary to the closest traffic predicted trajectory, otherwise, use the original quintic trajectory. 
\end{remark}

\begin{table}[htbp]
\caption{Frenet Algorithm Library}
\centering
\begin{tabular}{m{1.2cm}<{\centering}|m{2.3cm}|| m{1.2cm}<{\centering}|m{2.3cm}}
\Xhline{1.2pt}
\textbf{Algorithm ID}& \makecell[c]{\textbf{Description} }& \textbf{Algorithm ID} & \makecell[c]{\textbf{Description}}\\
\Xhline{1.2pt}
 \makecell[c]{\textit{QF-V} \\ (``Speed'')} & Quintic Frenet, only speed objective, tracking $v_d$. & \textit{EMATO-FV}  &  Optimize a homotopic trajectory to \textit{QF-V }. \\  
\hline
 \makecell[c]{\textit{QF-M} \\ (``Mixed'')} & Mixed with speed, jerk, and energy efficiency objectives.    & \textit{EMATO-FM} &  Optimize a homotopic trajectory to \textit{QF-M}. \\ 
\hline
 \makecell[c]{\textit{QF-E} \\ (``Energy'')}& Only energy efficiency objective. & \textit{EMATO-FE} & Optimize a homotopic trajectory to \textit{QF-E}. \\ 
\Xhline{1.2pt}

\end{tabular}

\label{tab:frenet}
\end{table}

%% file: sec4.tex
\section{Experiments}
\label{sec:experiments}
Extensive quantitative and ablation studies are conducted in ACC and Frenet scenarios with the abovementioned ``Truck'' model and a ``Sedan'' model from literature~\cite{down_slope}~(the original model is remapped w.r.t. $v$ and $a_t$, instead of $a_v$). In ACC Scenario, a one-lane car-following driving condition is simulated with the leading car driving in different standard cycles, they are,  highway cycles: ``HWFET''\cite{hwfet}, ``INDIA\_HWY''\cite{advisor}, urban cycles: 
  ``NYCC''\cite{nycc_cycle}, ``MANHATTAN''\cite{manhattan_cycle}, ``EUDC''\cite{EUDC}, ``NYC\_TRUCK'', ``INDIA\_URBAN''\cite{advisor}; In Frenet scenario, a three-lane road with traffic flow at 50, 56, 60 km/h is simulated; Planning and prediction horizon is fixed as $\Delta T = 5$s, time resolution $dt=0.1$s; NLP in EMATO is solved by Interior Point Optimizer~(IPOPT)~\cite{ipopt}, an efficient NLP algorithm with a CasADi~\cite{casadi} symbolic Python interface. Code and all parameters used are attached in ~\cite{emato_website}.

  \subsection{Quantative Study}
  In the quantitative study, we compare four algorithms in Table~\ref{tab:acc} in the ACC scenario, and six algorithms in Table~\ref{tab:frenet} for the Frenet Scenario, with metrics average speed, average jerk (absolute value), and fuel efficiency to respectively evaluate time efficiency, comfort, and energy efficiency (metrics in the Frenet scenario are compared on road coordinate $s$ instead of path coordinate $l$ for fairness). For a more efficient way to show the ACC results, 7 driving cycles are combined into a big cycle (leading car total travel distance as 41059.28m with a travel time of 4492.4s), hereby the total set number is reduced from $2~\text{(cars)} \times 3~\text{(road slope profiles)} \times 4~\text{(algorithms)} \times 7~\text{(cycles)} = 168$ to $24$; And similarly, $2\times3\times6=36$ results are obtained in 2200m-fixed-distance Frenet scenario tests. 
  
\begin{figure}
    \centering
    \includegraphics[width=\linewidth]{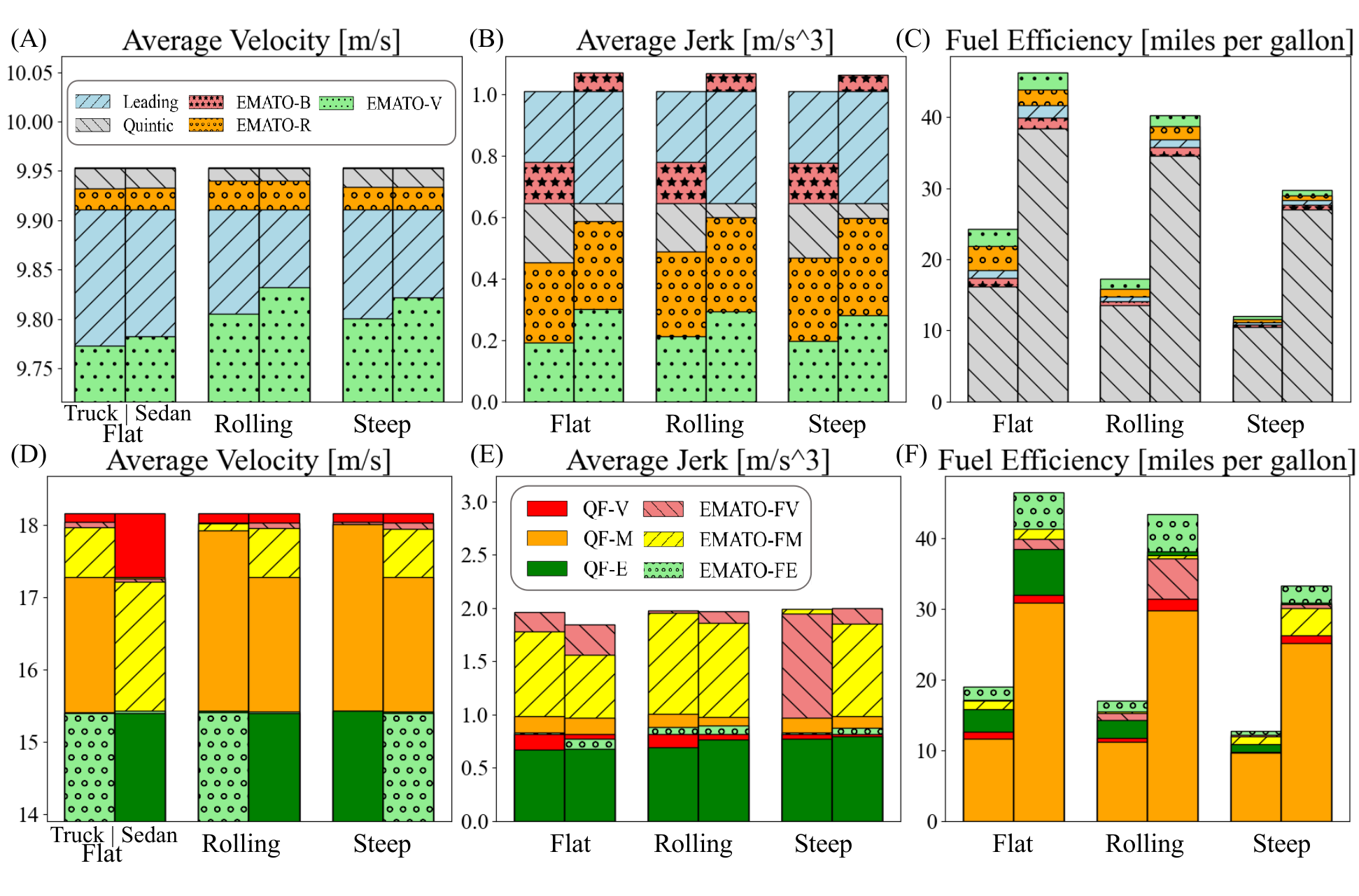}  
    \caption{(A-C) Evaluations for the ACC scenario; (D-F) Results for the Frenet scenario. EMATO improves fuel efficiency over the quintic polynomial trajectory interpolation method.}
\label{fig:quant}
\end{figure}

   \textbf{ACC results} are shown in Fig.~\ref{fig:quant}~(A-C), where all methods can track the leading car velocity well.  With end-state constraints becoming tighter, the average speed increases but fuel efficiency decreases. However, even tightly bounded \textit{EMATO-B} with quintic start-ends can get a 2.42\% to 7.02\% miles per gallon improvement as Table~\ref{tab:acc_results} shows. Moreover, in the \textbf{Frenet scenario}, \textit{QF-V} tracks a $v_d = 70 $km/h, hence performs the highest speed, while \textit{QF-M} and \textit{QF-E} have a lower speed. Since to some degree, EMATO produces a trajectory with PnG features over a quintic trajectory as Sec.~\ref{sec:png} illustrates, it increases the average jerk value. EMATO method, however, significantly improves the energy efficiency over the quintic methods as Table~\ref{tab:frenet_results} and Fig.~\ref{fig:quant}~(D-F) shows; The average solving time of EMATO is 0.0104s in an Intel i7 Ubuntu device, and is 0.0396s in an Nvidia Xavier device, which is considered online executable.

\begin{remark}
Over quintic trajectories, tightly bounded \textit{EMATO} homotopic trajectories, i.e., \textit{EMATO-B, EMATO-F(V,M,E)} demonstrate the trade-off between the jerk and the energy term, the trade-off is also mentioned in study~\cite{ocp_scp}.

\end{remark}

\begin{table}[h]
\caption{ACC Energy Efficiency Improvement}
\centering
\begin{tabular}{m{0.8cm}<{\centering}| m{0.8cm}<{\centering} |m{1.6cm}<{\centering}| m{1.6cm}<{\centering}| m{1.6cm}<{\centering}}
\Xhline{1.2pt}
\multirow{2}{*}[-0.5em]{\textbf{Slope}} & \multirow{2}{*}[-0.5em]{\textbf{Vehicle}} & \multicolumn{3}{c}{\textbf{Improvement (\%)}} \\
\cline{3-5}
 &  & \textbf{EMATO-B vs Quintic} & \textbf{EMATO-R vs Quintic} & \textbf{EMATO-V vs Quintic} \\
\Xhline{1.2pt}
\multirow{2}{*}[-0.5em]{Flat} & Truck & 7.02 & 35.11 & 49.97 \\
 & Sedan & 3.91 & 14.42 & 20.59 \\
\hline
\multirow{2}{*}[-0.5em]{Rolling} & Truck & 4.70 & 17.27 & 27.62 \\
 & Sedan & 3.48 & 12.32 & 16.65 \\
\hline
\multirow{2}{*}[-0.5em]{Steep} & Truck & 3.41 & 11.42 & 15.13 \\
 & Sedan & 2.42 & 7.26 & 10.19 \\
\Xhline{1.2pt}
\end{tabular}
\label{tab:acc_results}
\end{table}

\begin{table}[h]
\caption{Frenet Energy Efficiency Improvement}
\centering
\begin{tabular}{m{0.8cm}<{\centering}| m{0.8cm}<{\centering} |m{1.6cm}<{\centering}| m{1.6cm}<{\centering}| m{1.6cm}<{\centering}}
\Xhline{1.2pt}
\multirow{2}{*}[-0.5em]{\textbf{Slope}} & \multirow{2}{*}[-0.5em]{\textbf{Vehicle}} & \multicolumn{3}{c}{\textbf{Improvement (\%)}} \\
\cline{3-5}
 &  & \textbf{EMATO-FV vs QF-V} & \textbf{EMATO-FM vs QF-M} & \textbf{EMATO-FE vs QF-E} \\
\Xhline{1.2pt}
\multirow{2}{*}[-0.5em]{Flat} & Truck & 36.01 & 45.35 & 27.12 \\
 & Sedan & 24.58 & 33.94 & 25.99 \\
\hline
\multirow{2}{*}[-0.5em]{Rolling} & Truck & 30.02 & 37.69 & 24.91 \\
 & Sedan & 18.31 & 26.50 & 18.10 \\
\hline
\multirow{2}{*}[-0.5em]{Steep} & Truck & 25.06 & 24.63 & 18.54 \\
 & Sedan & 16.82 & 19.66 & 9.91 \\
\Xhline{1.2pt}
\end{tabular}
\label{tab:frenet_results}
\end{table}

\subsection{Ablation Study}
In the ablation study, 4 different objective functions with different weights in EMATO  are shaped to demonstrate the contribution of the fuel efficiency term. We use the ``HWFET'' cycle (leading car total travel distance as 16439.38 in 737.9s) in the ACC scenario with the truck model, \textit{EMATO-R} and three slope profiles, to do the ablation study. Weights $\mathbf{w}$ sets for $J$ with different objectives are given as $[[0]_{1\times3}, 1.16, 0]$ (``Jerk'' objective) , $[0, [9.51]_{1\times2}, 1.16, 0]$ (``General''),  $[[0]_{1\times4}, 1]$ (energy ``Efficiency''), $[0,[9.51]_{1\times2},1.16, 38.91]$ (``Holistic''). As Table~\ref{tab:ablation} illustrates, objectives of combining jerk and acceleration in the ``General'' setting could have better jerk and fuel efficiency than ``Jerk'', where the acceleration (general energy) term for energy-saving is a common setting for the energy-model-free methods~\cite{frenet,darpa_motion}. However, it is outperformed by the ``Efficiency'' setting which solely considers the energy efficiency term. The increased jerk in the ``Efficiency'' method is optimized by a comprehensive setting ``Holistic'' considering all jerk, acceleration, and energy efficiency terms. Because of the improvement in both jerk and efficiency terms, the ``Holistic'' setting is applied to all the tests in the quantitative study. 

\begin{table}[h]
    \caption{HWFET Ablation Study}
    \centering
    \begin{tabular}{
    >{\centering\arraybackslash}m{1cm}
    |>{\centering\arraybackslash}m{2.0cm}
    |>{\centering\arraybackslash}m{1cm}
    |>{\centering\arraybackslash}m{1cm}
    |>{\centering\arraybackslash}m{1.5cm}}
        \Xhline{1.2pt}
        \textbf{Slope Type} & \textbf{Weights $\mathbf{w}$}    & \textbf{Average Velocity} [m/s] & \textbf{Average Jerk} $[\text{m/s}^3]^2$ & \textbf{Fuel Efficiency} [mpg] \\
        \Xhline{1.2pt}
        \multirow{4}{*}{Flat} 
        & Jerk      & 22.17 & 0.08844 & 13.37 \\
        & General   & 22.17 & 0.02439 & 24.97 \\
        & Efficiency & 22.14 & 1.27129 & 26.36 \\
        \rowcolor{lime!20} 
        & Holistic  & 22.15 & \textbf{0.02228} & \textbf{26.54} \\
        \hline
        \multirow{4}{*}{Rolling} 
        & Jerk      & 22.12 & 0.08273 & 12.39 \\
        & General   & 22.16 & 0.03653 & 18.43 \\
        & Efficiency & 22.15 & 1.37754 & 18.72 \\
        \rowcolor{lime!20} 
        & Holistic  & 22.16 & \textbf{0.03609} & \textbf{19.22} \\
        \hline
        \multirow{4}{*}{Steep} 
        & Jerk      & 22.12 & 0.08180 & 10.02 \\
        & General   & 22.17 & \textbf{0.02618} & 13.43 \\
        & Efficiency & 22.16 & 1.36591 & \textbf{13.65} \\
        \rowcolor{lime!20} 
        & Holistic  & 22.17 & 0.02382 & 13.53 \\
        \Xhline{1.2pt}
    \end{tabular}

    \label{tab:ablation}
\end{table}

%% file: conclusion.tex
\section{Conclusion and Future Work}
\label{sec:conclusion}

This study introduces an energy-model-aware trajectory optimization method EMATO, that is naturally compatible with commonly used polynomial trajectories. By incorporating a precise energy model directly into the trajectory planning, a significant energy efficiency increase over polynomial methods by 2.42\% to 49.97\% is observed in different driving conditions for both a tested sedan and a truck, which potentially has an economical impact over billions of US dollars~\cite{truck_mileage}.  Besides the energy optimization on trajectory planning, in the future, the authors plan to research the energy problem introduced by behavior planning and decision-making. We also expect to integrate EMATO into Autoware~\cite{autoware} and validate the method in our real vehicles HydraU and HydraD~\cite{thecarlab}.